%% file: main.tex
\documentclass[10pt,twocolumn,letterpaper]{article}
\usepackage{times} 
\usepackage{authblk} 
\usepackage[accsupp]{axessibility}  

\usepackage{cvpr}              


\usepackage{algorithm}
\usepackage{algorithmic}
\usepackage{amssymb}   
\usepackage{amsmath} 
\usepackage[pagebackref,breaklinks,colorlinks,allcolors=cvprblue]{hyperref}
\usepackage[capitalize,noabbrev]{cleveref}
\usepackage{graphicx} 
\usepackage{multirow}
\usepackage{booktabs} 
\usepackage{xcolor}
\usepackage{amssymb}
\graphicspath{{CVPR26_EWC_lx/figure/}}
\usepackage{newfloat}
\usepackage{listings}
\DeclareCaptionStyle{ruled}{labelfont=normalfont,labelsep=colon,strut=off} 
\floatstyle{ruled}
\newfloat{listing}{tb}{lst}{}
\floatname{listing}{Listing}

\newcommand{\lx}[1]{\textcolor{black}{#1}}
\newcommand{\cxb}[1]{\textcolor{black}{#1}}

\definecolor{cvprblue}{rgb}{0.21,0.49,0.74}

\def\confName{CVPR}
\def\confYear{2026}

\title{Elastic Weight Consolidation Done Right for Continual Learning}

\author[1]{\vspace{-0.1cm}Xuan Liu}
\author[1,2,3]{Xiaobin Chang\thanks{indicates corresponding author.}}
\affil[1]{School of Artificial Intelligence, Sun Yat-sen University, China}
\affil[2]{Key Laboratory of Intelligent Assessment Technology for Sustainable Tourism, Ministry of Culture and Tourism, Sun Yat-sen University}
\affil[3]{Key Laboratory of Machine Intelligence and Advanced Computing, Ministry of Education, China}
\affil[ ]{\ttfamily\small liux687@mail2.sysu.edu.cn, changxb3@mail.sysu.edu.cn}

\begin{document}
\maketitle
\input{sec/0_abstract}    
\input{sec/1_intro}
\input{sec/2_relatedwork}

\input{sec/3_gradientanalysis}

\input{sec/4_logitsreversal}
\input{sec/5_experiments}

\input{sec/6_conclusion}


{
    \small
    \bibliographystyle{ieeenat_fullname}
    \bibliography{main}
}

\input{sec/X_suppl}

\end{document}

%% file: sec/0_abstract.tex
\begin{abstract}
Weight regularization methods in continual learning (CL) alleviate catastrophic forgetting by assessing and penalizing changes to important model weights.
Elastic Weight Consolidation (EWC) is a foundational and widely used approach within this framework that estimates weight importance based on gradients.
However, it has consistently shown suboptimal performance.
In this paper, we conduct a systematic analysis of importance estimation in EWC from a gradient-based perspective.
For the first time, we find that EWC's reliance on the Fisher Information Matrix (FIM) results in gradient vanishing and inaccurate importance estimation in certain scenarios.
Our analysis also reveals that Memory Aware Synapses (MAS), a variant of EWC, imposes unnecessary constraints on parameters irrelevant to prior tasks, termed the redundant protection.
Consequently, both EWC and its variants exhibit fundamental misalignments in estimating weight importance, leading to inferior performance.
To tackle these issues, we propose the Logits Reversal (LR) operation, a simple yet effective modification that rectifies EWC's importance estimation.
Specifically, reversing the logit values during the calculation of FIM can effectively prevent both gradient vanishing and redundant protection.
Extensive experiments across various CL tasks and datasets show that the proposed method significantly outperforms existing EWC and its variants. Therefore, we refer to it as EWC Done Right (EWC-DR)\footnote{Code is available at \url{https://github.com/scarlet0703/EWC-DR}.}.
\end{abstract}

%% file: sec/1_intro.tex
\section{Introduction}
\label{sec:intro}
\begin{figure}[!t]
    \centering
    \includegraphics[width=0.95\linewidth]{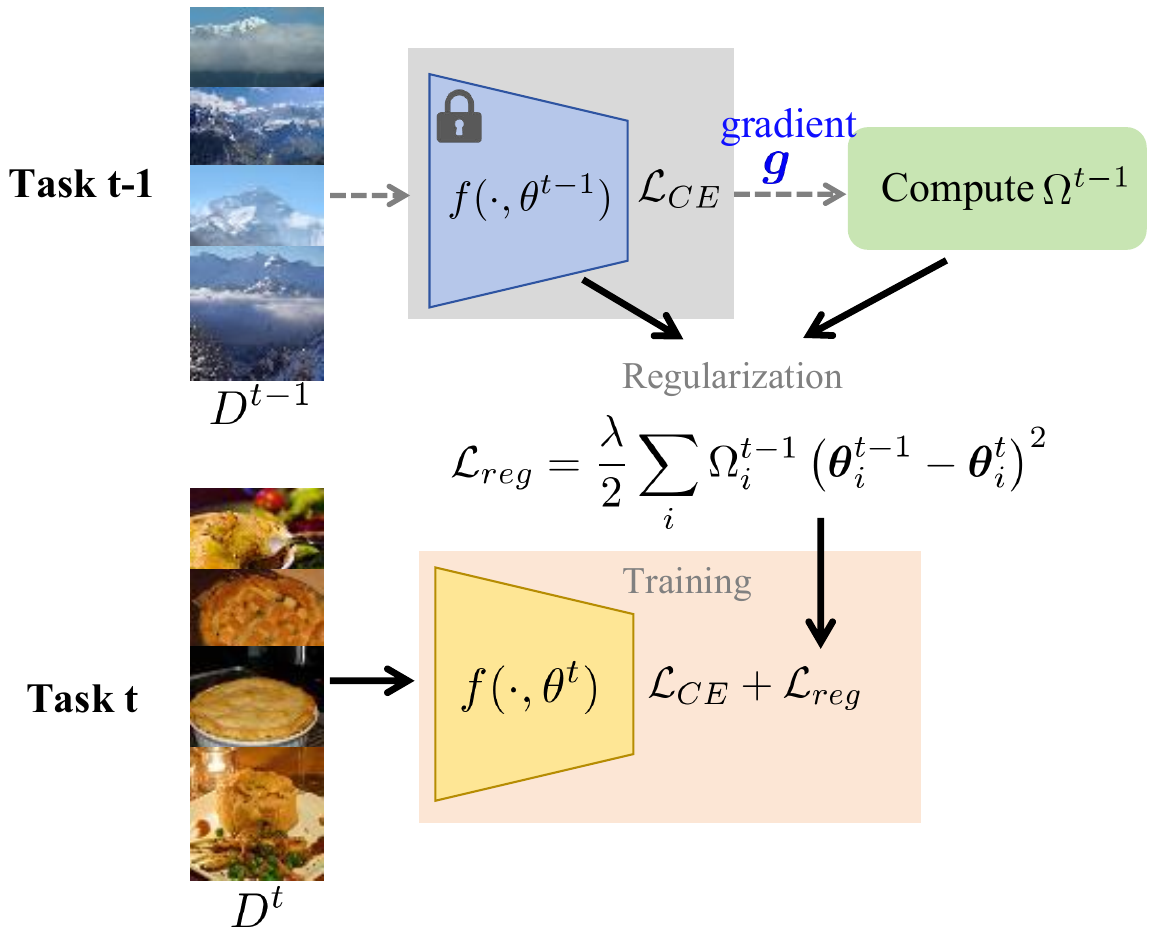}
    \caption{
        Overview of the EWC learning process.
        After training on task $t-1$, model weights $\theta^{t-1}$ are obtained. The gradient of the cross-entropy loss $\mathcal{L}_{CE}$ is computed using the dataset $\mathcal{D}^{t-1}$ with the model $f(\cdot,\theta^{t-1})$. These gradients estimate the weight importance matrix $\Omega^{t-1}$, but they are not used to update the weights. When learning a new task $t$, a regularization term $\mathcal{L}_{reg}$ based on $\Omega^{t-1}$ is added into the total loss to constrain changes and preserve important parameters in $\theta^{t-1}$.
    }
    \label{fig:ewc_process}
\end{figure}
Neural networks excel at computer vision tasks by training on samples from all tasks simultaneously. However, in continual learning (CL)~\cite{wang2024comprehensive,chen2018lifelong}, tasks are learned sequentially, with the network accessing data from only one task at a time. This leads to the challenge of catastrophic forgetting~\cite{mccloskey1989catastrophic}, where the network often forgets previously learned tasks when adapting to new ones.
One approach to handle this issue is to store a small subset of training data from earlier tasks to prevent forgetting during new task training~\cite{smith2024adaptive, zhou2022model, douillard2020podnet, liu2021adaptive, hu2021distilling}.
\cxb{However, these replay-based methods require storing and replaying past data, leading to issues like memory constraints, privacy concerns, and sampling bias~\cite{chen2023dynamic,babakniya2023data}.}
Alternatively, the network can be regularized to retain important weights from previous tasks without \cxb{exemplar} storage~\cite{rongali2020continual, kirkpatrick2017overcoming, aljundi2018memory},
\cxb{thereby eliminating memory overhead, avoiding privacy concerns, and enabling more scalable deployment in various CL scenarios.}
Elastic Weight Consolidation (EWC)~\cite{kirkpatrick2017overcoming} is a classic method that belongs to the second approach.

\lx{EWC remains relevant as a fundamental and widely adopted CL method for various tasks}, such as image classification~\cite{kim2023achieving,wang2021training,masana2022class,goswami2023fecam},  instruction tuning~\cite{chen2024coin,huai2025cl,guo2025hide,srinivasan2022climb}, and object detection~\cite{liu2020incdet,wang2021wanderlust,cermelli2022modeling}, due to its simplicity and rehearsal-free nature.
As shown in \cref{fig:ewc_process}, EWC assesses the importance of network weights for previously learned tasks by analyzing gradients and recording these importance scores. During training on new tasks, it regularizes the network to prevent significant changes in crucial weights, with the constraint strength determined by each weight's importance.
\cxb{However, EWC consistently demonstrates suboptimal performance~\cite{kim2023achieving,liu2018rotate,schwarz2018progress}.}
To examine EWC and its variants, we propose a novel gradient-based analysis framework.
Detailed analysis indicates that when the network achieves high confidence in accurate predictions, EWC produces a low-magnitude Fisher Information Matrix (FIM) due to the vanishing gradients. Consequently, EWC struggles to retain crucial parameter information, resulting in poor CL performance and significant catastrophic forgetting.
Applying this analysis to Memory Aware Synapses (MAS)~\cite{aljundi2018memory}, a variant of EWC, 
reveals that MAS can mitigate the gradient vanishing issue mentioned earlier, but introduces redundant protection by over-constraining parameters not related to previous tasks.
To resolve the weight importance misalignment, we introduce Logits Reversal (LR), an operation that simply reverses the logits during the computation of the gradient-based importance matrix.
This approach substantially improves the performance of vanilla EWC, which we refer to as EWC Done Right (EWC-DR).
The main contributions are summarized as follows:
\begin{itemize}
    \item We analyze EWC and its variants \cxb{from} a gradient-based \cxb{perspective}, offering new insights for developing more reliable and effective regularization-based CL algorithms.
    \item The analysis shows that EWC and its variants exhibit weight importance misalignments: EWC suffers from gradient vanishing, while MAS experiences redundant protection, leading to inferior CL performance.
    \item We propose EWC-DR, an enhancement to vanilla EWC \cxb{by incorporating} the operation of logits reversal (LR) to better estimate weight importance and improve CL results.
\end{itemize}
\cxb{We demonstrate the improved effectiveness of the proposed EWC-DR across different continual learning tasks, including exemplar-free class-incremental learning and continual learning of multimodal instruction tuning.
EWC-DR significantly outperforms vanilla EWC and its variants, such as online EWC, MAS, and Synaptic Intelligence (SI).}

%% file: sec/2_relatedwork.tex
\section{Related Work}
\subsection{Continual Learning Scenarios}

\subsubsection{Class Incremental Learning} 
In Class Incremental Learning (CIL), the model is trained on a sequence of non-overlapping $T$ tasks $\{ 1, 2, \dots, \mathcal{T} \}$, where each new task introduces new classes to the output space. The number of available classes increases incrementally over time, making this setting more challenging and realistic. The data distribution of task $t$, denoted as $\mathcal{D}^{t}$, is fixed but unknown in advance, while all tasks share the same output space $\mathcal{Y}$. Specifically, $\mathcal{D}^{t}$ consists of $N^{t}$ labeled examples $\{ (\boldsymbol{x}^{t}_{n}, \boldsymbol{y}^{t}_{n}) \}_{n=1}^{N^{t}}$, where the new labels $\boldsymbol{y}^{t}_{j}$ belong to a task-specific subset $\mathcal{Y}^{t} \subset \mathcal{Y}$, and no label overlap exists between tasks, i.e., $ \bigcup_{j=1}^{t-1} \mathcal{Y}^{j} \cap \mathcal{Y}^{t} = \emptyset$. At the end of task $t$, the total number of seen classes becomes $\sum_{j=1}^{t} |\mathcal{Y}^{j}|$. Depending on whether past samples (exemplars) are stored and reused during training, CIL methods are generally categorized into \emph{exemplar-based} (EBCIL)~\cite{zhou2022model,rebuffi2017icarl,iscen2020memory,zhao2021memory,chaudhry2018riemannian} and \emph{exemplar-free} (EFCIL)~\cite{zhu2021prototype,petit2023fetril,zhu2021class,zhu2022self,yu2020semantic} approaches.
\cxb{In CIL literature, exemplar-based methods typically outperform exemplar-free ones.
However, in constrained scenarios involving privacy and limited resources, storing previous task data samples is prohibited, making EFCIL more challenging.
Thus, direct comparisons between these two settings can be misleading.}

\subsubsection{Multimodal Continual Instruction Tuning}
In Multimodal Continual Instruction Tuning (MCIT), the goal is to continually adapt a multimodal model to new data while preserving previously acquired knowledge~\cite{baltruvsaitis2018multimodal,srinivasan2022climb}.
Consider a sequence of datasets $\{\mathcal{D}^1, \ldots, \mathcal{D}^{\mathcal{T}}\}$, where each $\mathcal{D}^{t} = \{(\boldsymbol{x}_n^{\mathrm{img}}, \boldsymbol{x}_n^{\mathrm{text}}, \boldsymbol{y}_n)\}_{n=1}^{N^t}$ contains $N^t$ triplets, consisting of an image, a text instruction, and the corresponding answer.  
This continual learning paradigm closely resembles how humans continuously learn from diverse modalities in real-world scenarios~\cite{mroczko2016perception}.
Recently, the weight regularization method EWC has been widely used as a baseline for comparison in MCIT to evaluate its effectiveness in mitigating catastrophic forgetting~\cite{chen2024coin,liu2025c,srinivasan2022climb,guo2025hide,huai2025cl}.

\subsection{Weight Regularization-based Methods}
Elastic Weight Consolidation (EWC)~\cite{kirkpatrick2017overcoming} is a classic weight regularization-based method for continual learning.
It estimates the importance of each parameter after training on a task. An importance matrix $\Omega$ is maintained, where a larger $\Omega_i$ indicates that parameter $\theta_i$ is more critical. 
To prevent forgetting of the previous task $t-1$ during training on a new task $t$, EWC penalizes changes to important parameters by adding a regularization loss
\begin{equation}
\mathcal{L}_{reg} = \frac{\lambda}{2} \sum_i \Omega_i^{t-1} \left( \theta_i^{t-1} - \theta_i^{t} \right)^2,
\label{ewc_regloss}
\end{equation}
where $\theta_i^{t-1}$ is the $i$-th parameter after training on the previous task data $\mathcal{D}^{t-1}$, and $\left( \theta_i^{t} - \theta_i^{t-1} \right)^2$ measures the deviation from the previously learned parameter.

The importance $\Omega$ is calculated based on an approximation of the Fisher Information Matrix (FIM), and many studies~\cite{kim2023achieving,liu2018rotate,schwarz2018progress} have highlighted that such approximations can be inaccurate in practice.
To address this, Online EWC~\cite{schwarz2018progress} proposes to accumulate the importance weights in an online manner across tasks, introducing a decay factor to gradually forget outdated information and reduce memory overhead. 
Unlike EWC, which uses FIM to assess parameter importance, Synaptic Intelligence (SI)~\cite{zenke2017continual} computes importance online by tracking each parameter's contribution to loss reduction during training.
Memory Aware Synapses (MAS)~\cite{aljundi2018memory} measures importance based on the sensitivity of the learned output function to each parameter. Parameters that cause larger changes in the output when perturbed are assigned higher importance values.
However, EWC and its variants lack theoretical analyses and do not identify the fundamental cause of their performance degradation.

%% file: sec/3_gradientanalysis.tex
\section{Gradient-based Analysis of EWCs}
In this section, motivated by the observation that weight importance in EWC-based methods is closely related to gradient values, we propose a gradient-based analytical framework to theoretically and empirically examine existing approaches to importance estimation.
Specifically, our analysis focuses on the fully connected (FC) layer, as its parameters are directly associated with classification outcomes, making them more interpretable.
Furthermore, according to the chain rule, gradients from the FC layer are backpropagated through the network, thereby affecting the updates to preceding parameters.
Thus, we focus on the importance of FC layer weights in our analysis for simplicity and clarity.
In continual learning for classification, the goal is to learn $T$ tasks sequentially while retaining knowledge from earlier tasks.
For task $t-1$, let the training dataset be denoted as $\mathcal{D}^{t-1} = \{ (\boldsymbol{x}_n^{t-1}, \boldsymbol{y}_n^{t-1}) \}_{n=1}^{N^{t-1}}$, where $N^{t-1}$ is the number of samples and $C^{t-1}$ denotes the number of classes encountered up to task $t-1$.
After training on $\mathcal{D}^{t-1}$, we obtain a model parameterized by $\theta^{t-1}= \{\theta^{t-1}_i\}_{i}$. Weight regularization methods estimate the importance of each parameter ${\theta}_i^{t-1}$ by computing an importance weight $\Omega_i^{t-1}$. A larger value of $\Omega^{t-1}_i$ indicates that parameter $i$ is more important for task $t-1$.
When learning a new task $t$, importance values $\Omega^{t-1} = \{\Omega^{t-1}_i\}_{i}$ are used in a regularization loss (\cref{ewc_regloss}) to preserve important parameters in $\theta^{t-1}$ and mitigate catastrophic forgetting.

Considering the FC layer of the classification head,
where the input is $\boldsymbol{h}^{t-1} \in \mathbb{R}^{n}$, and the weight matrix is $w^{t-1} \in \mathbb{R}^{C \times n}$. The output logits are computed as $\boldsymbol{z}^{t-1} = w^{t-1} \boldsymbol{h}^{t-1} \in \mathbb{R}^{C}$. The weights $w^{t-1}= \{w^{t-1}_k\}_{k}$ are a subset of the entire network parameter set, $w^{t-1} \subseteq \theta^{t-1}$. 
The predicted probability for each class $k \in \{1, \ldots, C\}$ is then obtained via the softmax function
\begin{equation}
    p_k^{t-1} = \mathrm{Softmax}(\boldsymbol{z}^{t-1})_k = \frac{\exp(z^{t-1}_k)}{\sum_{j=1}^{C} \exp(z^{t-1}_j)}.
\label{softmax}
\end{equation}
The predicted class is given by the index with the largest probability, i.e., $\hat{y}^{t-1} = \arg\max_{k} p^{t-1}_k$.
In the following analysis, we focus on the calculation of the parameter importance $\Omega_{w}^{t-1}$ for task $t-1$, with particular attention to the FC layer weights $w^{t-1}$. Therefore, we will omit the superscript $t-1$ from the notation for simplicity.

\subsection{Gradient Vanishing in EWC}
In Elastic Weight Consolidation (EWC), the importance $\Omega_i^{EWC}$ of each model parameter is approximated via the
diagonal FIM
and defined as
\begin{equation}
\Omega_i^{EWC} = F_i = \mathbb{E}_{(\boldsymbol{x},\boldsymbol{y}) \sim p(\boldsymbol{x},\boldsymbol{y})} \left[ \left( \frac{\partial \log p(\boldsymbol{y} \mid \boldsymbol{x}, \theta)}{\partial \theta_i} \right)^2 \right],
\label{ewc_fim1}
\end{equation}
where $p(\boldsymbol{y} \mid \boldsymbol{x}, \theta)$ denotes the model's predictive distribution. 
Since the cross-entropy loss $\mathcal{L}_{CE}$ is the negative log-likelihood, under the assumption that $\theta$ is close to a local optimum, the FIM can equivalently be approximated by the squared gradients of the cross-entropy loss evaluated on the training data
\begin{equation}
\mathcal{L}_{CE}(\theta) = -\log p(\boldsymbol{y} \mid \boldsymbol{x}, \theta) = -\sum_k y_k \log(p_k),
\label{ewc_ce}
\end{equation}
\begin{equation}
F_i = \mathbb{E}_{(\boldsymbol{x},\boldsymbol{y})}\left[ \left( \frac{\partial \mathcal{L}_{CE}(\theta)}{\partial \theta_i} \right)^2 \right], \label{eq:ewc_fimgrad}
\end{equation}
where $y_k$ is the $k$-th value in the one-hot encoded label $\boldsymbol{y}$. Assuming the ground-truth class is $c$, we have
\begin{equation}
y_k = 
\begin{cases}
1, & \text{if } k = c, \\
0, & \text{if } k \neq c.
\end{cases}
\end{equation}
Consequently, \cxb{parameters with larger gradient variations are considered more critical for retaining knowledge acquired from the current task.}
The gradient of $\mathcal{L}_{CE}$ with respect to the $k$-th class of FC layer weight $w$ is given by
\begin{equation}
\frac{\partial \mathcal{L}_{\mathrm{CE}}}{\partial w_k} = 
\frac{\partial \mathcal{L}_{\mathrm{CE}}}{\partial z_k} \cdot 
\frac{\partial z_k}{\partial w_k}.
\end{equation}
Considering $k = c$, thus $y_c =1$, the derivative of $\mathcal{L}_{CE}$ with respect to $z_c$ is
\begin{equation}\label{eq:ewc_py_1}
\frac{\partial \mathcal{L}_{\mathrm{CE}}}{\partial z_c} = p_c - 1 = p_c - y_c.
\end{equation}
Otherwise ($k\neq c$), $y_k =0$ and $\frac{\partial \mathcal{L}_{\mathrm{CE}}}{\partial z_k}$ becomes
\begin{equation}\label{eq:ewc_py_2}
\frac{\partial \mathcal{L}_{\mathrm{CE}}}{\partial z_k} = p_k= p_k - y_k.
\end{equation}
Therefore, \cref{eq:ewc_py_1,eq:ewc_py_2} can be unified as
\begin{equation}\label{eq:ewc_py}
\frac{\partial \mathcal{L}_{\mathrm{CE}}}{\partial z_k} = p_k - y_k.
\end{equation}
Accordingly, the gradient with respect to $w_k$ is
\begin{equation}
\frac{\partial \mathcal{L}_{\mathrm{CE}}}{\partial w_k} = (p_k - y_k) \cdot \frac{\partial z_k}{\partial w_k}.
\end{equation}
The importance of the FC layer weight $w_k$ corresponding to the $k$-th class for EWC can be expressed as
\begin{equation}
\Omega_{w_k}^{\mathrm{EWC}} = F_{w_k} = \mathbb{E}\left[ (p_k - y_k )^2 \cdot (\frac{\partial z_k}{\partial w_k})^2 \right] .
\end{equation}

\begin{figure}[!t]
\centering
\begin{subfigure}{0.47\linewidth}
  \centering
  \includegraphics[width=\linewidth]{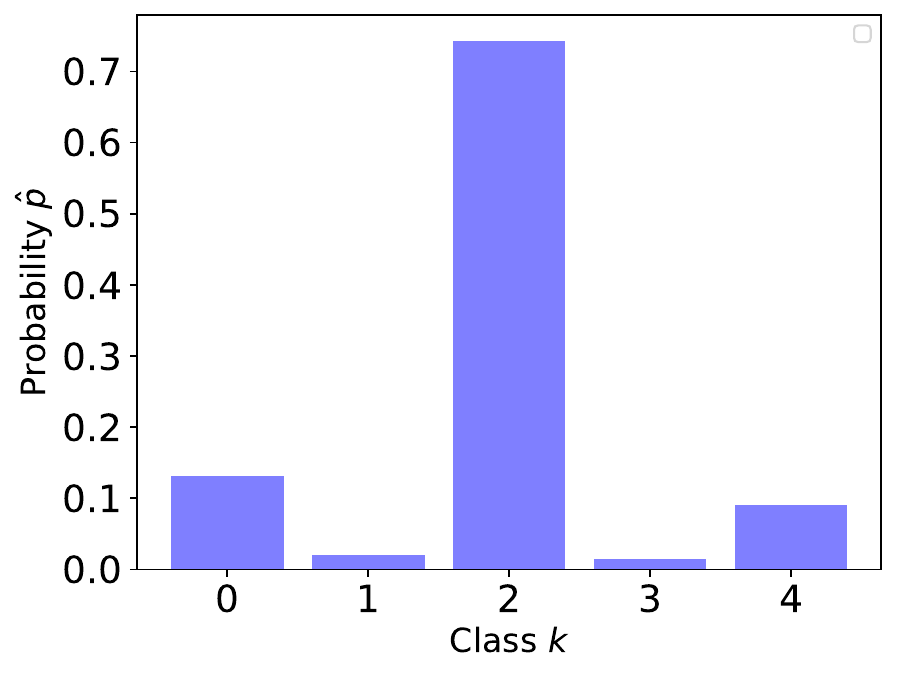}
  \caption{The predicted probability $p_k$ of $k$-th class. Ground truth at 2.}
  \label{fig:subfig1}
\end{subfigure}\hspace{2mm}
\begin{subfigure}{0.47\linewidth}
  \centering
  \includegraphics[width=\linewidth]{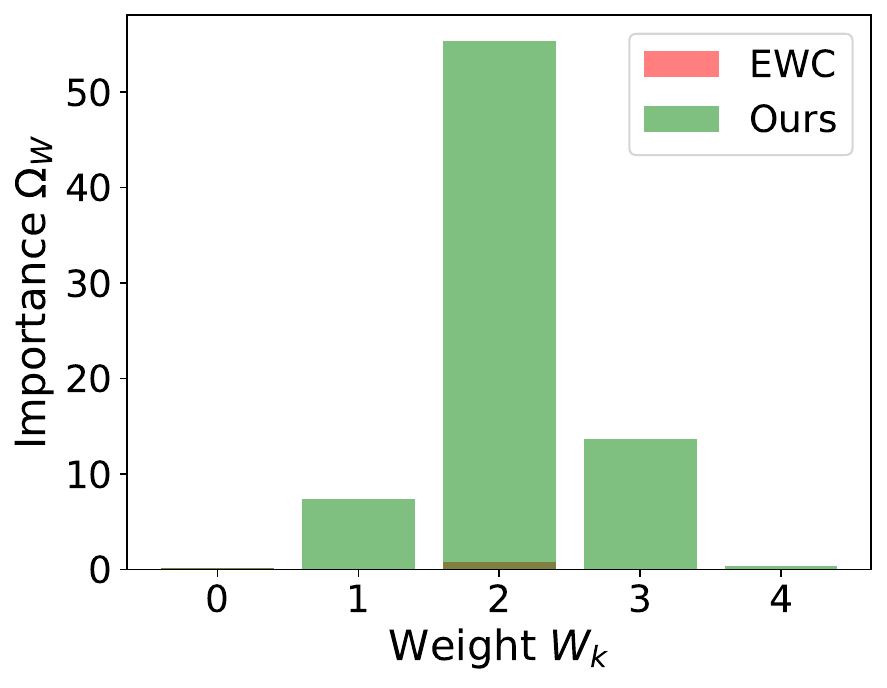}
  \caption{Importance of the FC layer weights, summed per class.}
  \label{fig:subfig2}
\end{subfigure}
\caption{A Case Study of Gradient Vanishing.}
\label{fig:gradient_vanishing}
\end{figure}

However, when the model makes the correct prediction at a high confidence, i.e., the predicted probability of \( p_c \) approaches 1, making the difference \( (p_c - 1) \) close to 0, which we refer to as \textbf{gradient vanishing}.
Therefore, the FIM value \( F_{w_c} \) for the final FC layer weight $w_c$, crucial to correct classification, is underestimated, failing to properly represent its significance.
The FIM value \( F_{w_k} \) for weight $w_k$, $k \neq c$, also experiences gradient vanishing due to the corresponding \( p_k \) approaches 0.
This problem can be illustrated with the case in \cref{fig:gradient_vanishing}, the predicted probability $p_2$ of a sample is close to 1 (\cref{fig:gradient_vanishing}(a)), but the importance scores $\Omega_{w_2}^{\mathrm{EWC}}$ for the FC weight parameters across all classes are very small (red bars in \cref{fig:gradient_vanishing}(b)).
Since the FIM is computed from the trained model using the training data at the end of each task,
\cxb{this issue frequently arises,}
ultimately affecting EWC's ability to accurately retain important knowledge.

\subsection{Redundant Protection in MAS}

\begin{figure}[!t]
\centering
\begin{subfigure}{0.47\linewidth}
  \centering
  \includegraphics[width=\linewidth]{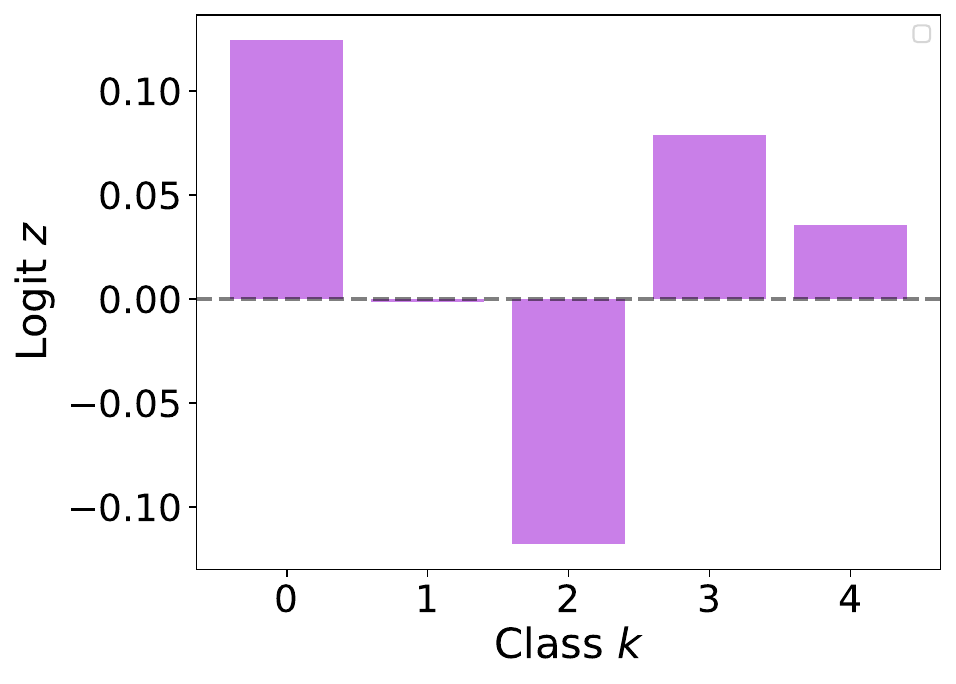}
  \caption{The logit $z_k$ of $k$-th class. Ground truth at 0.}
  \label{fig:subfig3}
\end{subfigure}\hspace{2mm}
\begin{subfigure}{0.47\linewidth}
  \centering
  \includegraphics[width=\linewidth]{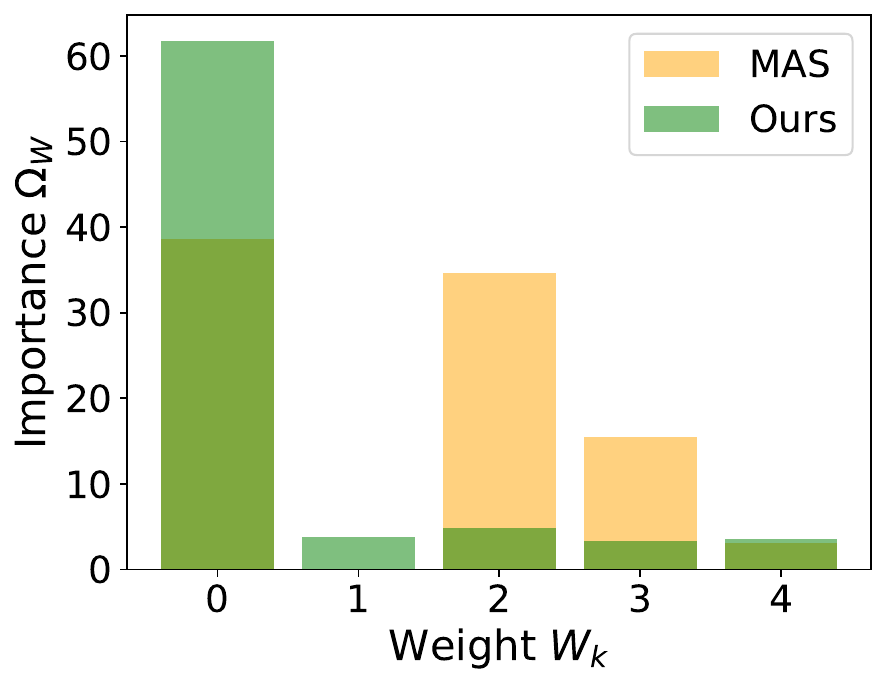}
  \caption{Importance of the FC layer weights, summed per class.}
  \label{fig:subfig4}
\end{subfigure}
\caption{A Case Study of Redundant Protection.}
\label{fig:redundant_protection}
\end{figure}
Memory Aware Synapses (MAS) is a variant of EWC. Instead of relying on the FIM, MAS quantifies parameter importance by measuring \cxb{how the $\ell_2$ norm of the network output changes with small parameter changes.}
For a neural network with parameters $\theta$, and given input $x$, let the output be $f(\boldsymbol{x}; \theta)$. The loss $\mathcal{L}_2$ and the importance $\Omega_i^{MAS}$ are defined as
\begin{equation}
\mathcal{L}_2(\theta) =  \| f(\boldsymbol{x}; \theta) \|_2 ,
\end{equation}

\begin{equation}
\Omega_i^{MAS} = \left\| \frac{\partial \mathcal{L}_2}{\partial \theta_i} \right\|_1.
\end{equation}
For the weight matrix $w$ in an FC layer, the logits $\boldsymbol{z}$ are the output. The corresponding gradient of the $\ell_2$ norm loss with respect to the $k$-th class FC layer weight $w_k$ can be computed as
\begin{equation}
\frac{\partial \mathcal{L}_2}{\partial w_k} = \frac{\partial \mathcal{L}_2}{\partial z_k} \cdot \frac{\partial z_k}{\partial w_k}.
\end{equation}
For each class $k$, the gradient with respect to the $k$-th logit $z_k$ is
\begin{equation}
\frac{\partial \mathcal{L}_2}{\partial z_k} = \frac{z_{k}}{\|\boldsymbol{z}\|_2},
\end{equation}
Thus, the gradient of the $\ell_2$ norm loss is simply the logits value itself.
Therefore, the gradient with respect to $w_k$ is
\begin{equation}
\frac{\partial \mathcal{L}_2}{\partial w_k} = \frac{z_k}{\|\boldsymbol{z}\|_2} \cdot \frac{\partial z_k}{\partial w_k}.
\end{equation}

The importance of the FC layer weight $w_k$ corresponding to the $k$-th class for MAS can be expressed as
\begin{equation}
\Omega_{w_k}^{\mathrm{MAS}} =  \left\| \frac{\partial \mathcal{L}_2}{\partial w_k} \right\|_1 = \frac{|z_{k}|}{\|\boldsymbol{z}\|_2} \cdot |\frac{\partial z_{k}}{\partial w_k}|.
\end{equation}

However, since MAS computes weight importance using the $\ell_2$ norm without a softmax operation, the logits remain unbounded. Consequently, when 
\cxb{a} large negative logit $z_k$ appear, MAS may assign disproportionately high importance to parameters associated with these branches, resulting in \textbf{redundant protection}.
This issue is illustrated in \cref{fig:redundant_protection}, where logit $z_2$ is a large negative value (corresponding to a very small predicted probability $p_2$), yet the importance score $\Omega_{w_2}^{\mathrm{MAS}}$ for the associated FC weight parameters are very high, as the last yellow bar \cref{fig:redundant_protection}(b) indicates.
Since such extreme negative logits have \cxb{little} impact on the output probability, preserving these parameters is unnecessary and can even hinder the model's plasticity in learning new tasks.

%% file: sec/4_logitsreversal.tex
\section{Logits Reversal}
\cxb{Based on the analysis provided, we propose a new method for effectively implementing EWC}.
This method introduces a logits reversal (LR) operation during the computation of the Fisher Information Matrix (FIM). 
\cxb{By negating the logits in the softmax output, we highlight key parameters that enhance the accuracy of class predictions.}

Specifically, during the computation of weight importance, we negate each logit \( z_k \) to obtain $\tilde{z}_k$
\begin{equation}
\tilde{z}_k = -z_k.
\end{equation}
Under this transformation, the new softmax output becomes
\begin{equation}
\tilde{p}_k = \frac{e^{\tilde{z}_k}}{\sum_j e^{\tilde{z}_j}} = \frac{e^{-z_k}}{\sum_j e^{-z_j}}.
\end{equation}
The corresponding cross-entropy loss with the LR softmax output $\tilde{p}_k$ is defined as
\begin{equation}
\tilde{\mathcal{L}}_{CE} = -\sum_k y_k \log(\tilde{p}_k).
\end{equation}
For the correct class $k = c$, the derivative of $\tilde{\mathcal{L}}_{CE}$ with respect to $k$-th class of the FC layer weight $w$ is
\begin{equation}\label{eq:ewcdr_py1}
\frac{\partial \tilde{\mathcal{L}}_{CE}}{\partial z_c} = 1 - \tilde{p}_c = y_c - \tilde{p}_c.
\end{equation}
whereas for incorrect classes ($k \neq c$),
\begin{equation}\label{eq:ewcdr_py2}
\frac{\partial \tilde{\mathcal{L}}_{CE}}{\partial z_k} =  - \tilde{p}_k = y_k - \tilde{p}_k.
\end{equation}
By unifying~\cref{eq:ewcdr_py1} and~\cref{eq:ewcdr_py2}, the importance of the FC layer weight $w$ corresponding to the $k$-th class is given by
\begin{equation}
\Omega_{w_k}^{LR} =\mathbb{E}\left[ (y_k - \tilde{p}_k )^2 \cdot (\frac{\partial \tilde{z}_k}{\partial w_k)})^2 \right].
\end{equation}

To understand the behavior of $\tilde{p}_k$ under LR, we analyze how the output probability responds to changes in its associated logit
\begin{equation}
\frac{\partial \tilde{p}_k}{\partial z_k} = \frac{e^{-z_k} \left( e^{-z_k} - \sum_j e^{-z_j} \right)}{\left( \sum_j e^{-z_j} \right)^2}.
\end{equation}
Given that $e^{-z_k} > 0$ and $e^{-z_k} - \sum_j e^{-z_j} < 0$ (unless $e^{-z_k}$ dominates the sum, which is rarely the case), it follows that
\begin{equation}
\frac{\partial \tilde{p}_k}{\partial z_k} < 0.
\end{equation}
Thus, after LR, increasing $z_k$ leads to a decrease in $\tilde{p}_k$, which is the opposite of the original softmax behavior.
As $z_c$ increases, $\tilde{p}_c$ for the correct class decreases, making $1 - \tilde{p}_c$ larger and thus increasing $\Omega_{w_c}^{\mathrm{LR}}$. At the same time, $\tilde{p}_k$ for incorrect classes remains small, resulting in negligible gradient contributions from these classes. Therefore, when computing the importance $\Omega_{w_k}^{\mathrm{LR}}$ under LR, the dominant contribution to the parameter importance arises from the correct class.

As shown in~\cref{fig:gradient_vanishing}, our approach (green bars) assigns the highest importance to the correct class (class~2), even when $z_2$ is close to 1. In high-confidence cases, LR also maintains
gradient magnitude, allowing the FIM scores to more accurately reflect the importance of each weight.
\cref{fig:redundant_protection} demonstrates that our approach (green bars) prioritizes the ground-truth class (class 0) while MAS unnecessarily protects class 2.
LR can effectively avoid this redundancy by focusing on the parameters that truly impact accurate predictions.
Therefore, computing $\Omega_{w_k}^{\mathrm{LR}}$ with LR offers a more discriminative estimation of weight importance across tasks.
Statistics of the FC layer weight importance are provided in the next section.

%% file: sec/5_experiments.tex
\section{Experiments}
\label{sec::experiments}
In this section, we compare our method, EWC-DR, with EWC and its variants in \cxb{standard exemplar-free class-incremental learning (EFCIL) settings, following the big-start and equally split protocols.}
We also evaluate its performance in multimodal continual instruction tuning (MCIT).

\begin{table*}[t]
\centering
\caption{EFCIL performance on CIFAR-100. Reported results represent means across three independent trials.}
\begin{tabular}{ccccccccccccc}
\toprule
\multirow{3}{*}{method} & \multicolumn{6}{c}{Big start} & \multicolumn{6}{c}{Equally split} \\\cmidrule(r){2-7} \cmidrule(r){8-13}
& \multicolumn{2}{c}{T=5} & \multicolumn{2}{c}{T=10} & \multicolumn{2}{c}{T=20} & \multicolumn{2}{c}{T=5} & \multicolumn{2}{c}{T=10} & \multicolumn{2}{c}{T=20}\\ 
\cmidrule(r){2-3} \cmidrule(r){4-5} \cmidrule(r){6-7} \cmidrule(r){8-9} \cmidrule(r){10-11} \cmidrule(r){12-13}
& \multicolumn{1}{c}{$A_{last}$} & $A_{avg}$ & \multicolumn{1}{c}{$A_{last}$} & $A_{avg}$ & \multicolumn{1}{c}{$A_{last}$} &$A_{avg}$ & \multicolumn{1}{c}{$A_{last}$} & $A_{avg}$ & \multicolumn{1}{c}{$A_{last}$} & $A_{avg}$ & \multicolumn{1}{c}{$A_{last}$} &$A_{avg}$\\
\midrule
EWC & \multicolumn{1}{c}{} 14.61& 32.82& \multicolumn{1}{c}{} 11.53& 27.02& \multicolumn{1}{c}{} 6.20& 19.61& \multicolumn{1}{c}{} 17.80& 39.02& \multicolumn{1}{c}{} 10.62& 27.90& \multicolumn{1}{c}{} 5.48& 18.13\\
online EWC & \multicolumn{1}{c}{} 29.70& 45.65& \multicolumn{1}{c}{} 26.36& 40.36& \multicolumn{1}{c}{} 17.63& 31.71& \multicolumn{1}{c}{} 26.13& 43.97& \multicolumn{1}{c}{} 17.45& 33.04& \multicolumn{1}{c}{} 11.52& 25.49\\
SI & \multicolumn{1}{c}{} 19.26& 37.68& \multicolumn{1}{c}{} 11.33& 27.45& \multicolumn{1}{c}{} 5.62& 17.62& \multicolumn{1}{c}{} 13.10& 29.84& \multicolumn{1}{c}{} 9.91& 23.58& \multicolumn{1}{c}{} 7.98& 20.72\\
MAS & \multicolumn{1}{c}{} 35.37& 48.32& \multicolumn{1}{c}{} 29.09& 43.24& \multicolumn{1}{c}{} 19.98& 34.77& \multicolumn{1}{c}{} 29.45& 43.68& \multicolumn{1}{c}{} 23.41& 37.84& \multicolumn{1}{c}{} 13.24& 27.39\\
EWC-DR & \multicolumn{1}{c}{} \textbf{50.23}& \textbf{63.75}& \multicolumn{1}{c}{} \textbf{44.88}& \textbf{60.94}& \multicolumn{1}{c}{} \textbf{35.86}& \textbf{53.45}& \multicolumn{1}{c}{} \textbf{46.89}& \textbf{61.47}& \multicolumn{1}{c}{} \textbf{29.41}& \textbf{46.01}& \multicolumn{1}{c}{} \textbf{18.00}& \textbf{33.52}\\
\bottomrule
\end{tabular}
\label{tab:efcil_cifar}
\end{table*}

\begin{table*}[t]
\caption{EFCIL performance on ImageNet-Subset. Reported results represent means across three independent trials.}
\centering
\begin{tabular}{ccccccccccccc}
\toprule
\multirow{3}{*}{method} & \multicolumn{6}{c}{Big start} & \multicolumn{6}{c}{Equally split} \\\cmidrule(r){2-7} \cmidrule(r){8-13}
& \multicolumn{2}{c}{T=5} & \multicolumn{2}{c}{T=10} & \multicolumn{2}{c}{T=20} & \multicolumn{2}{c}{T=5} & \multicolumn{2}{c}{T=10} & \multicolumn{2}{c}{T=20}\\ 
\cmidrule(r){2-3} \cmidrule(r){4-5} \cmidrule(r){6-7} \cmidrule(r){8-9} \cmidrule(r){10-11} \cmidrule(r){12-13}
& \multicolumn{1}{c}{$A_{last}$} & $A_{avg}$ & \multicolumn{1}{c}{$A_{last}$} & $A_{avg}$ & \multicolumn{1}{c}{$A_{last}$} &$A_{avg}$ & \multicolumn{1}{c}{$A_{last}$} & $A_{avg}$ & \multicolumn{1}{c}{$A_{last}$} & $A_{avg}$ & \multicolumn{1}{c}{$A_{last}$} &$A_{avg}$\\
\midrule
EWC & \multicolumn{1}{c}{} 11.44& 26.57& \multicolumn{1}{c}{} 5.76& 15.52& \multicolumn{1}{c}{} 3.73& 9.86& \multicolumn{1}{c}{} 20.78& 42.37& \multicolumn{1}{c}{} 10.60& 28.75& \multicolumn{1}{c}{} 5.28& 18.74\\
online EWC & \multicolumn{1}{c}{} 23.56& 46.68& \multicolumn{1}{c}{} 16.12& 39.71& \multicolumn{1}{c}{} 13.00& 32.74& \multicolumn{1}{c}{} 26.64& 46.58& \multicolumn{1}{c}{} 12.78& 29.72& \multicolumn{1}{c}{} 6.34& 18.91\\
SI & \multicolumn{1}{c}{} 9.19& 24.56& \multicolumn{1}{c}{} 4.80& 13.66& \multicolumn{1}{c}{} 2.93& 8.30& \multicolumn{1}{c}{} 17.83& 40.19& \multicolumn{1}{c}{} 9.10& 26.44& \multicolumn{1}{c}{} 4.77& 16.77\\
MAS & \multicolumn{1}{c}{} 21.06& 42.59& \multicolumn{1}{c}{} 13.72& 31.17& \multicolumn{1}{c}{} 7.86& 22.44& \multicolumn{1}{c}{} 45.33& 62.99& \multicolumn{1}{c}{} 24.36& 44.16& \multicolumn{1}{c}{} 11.47& 26.83\\
EWC-DR & \multicolumn{1}{c}{} \textbf{66.18}& \textbf{76.00}& \multicolumn{1}{c}{} \textbf{58.94}& \textbf{70.99}& \multicolumn{1}{c}{} \textbf{43.43}& \textbf{59.59}& \multicolumn{1}{c}{} \textbf{54.80}& \textbf{68.46}& \multicolumn{1}{c}{} \textbf{35.02}& \textbf{54.55}& \multicolumn{1}{c}{} \textbf{17.17}& \textbf{34.68}\\
\bottomrule
\end{tabular}
\label{tab:efcil_imsub}
\end{table*}

\begin{table*}[t]
\caption{EFCIL performance on Tiny-ImageNet. Reported results represent means across three independent trials.}
\centering
\begin{tabular}{ccccccccccccc}
\toprule
\multirow{3}{*}{method} & \multicolumn{6}{c}{Big start} & \multicolumn{6}{c}{Equally split} \\\cmidrule(r){2-7} \cmidrule(r){8-13}
& \multicolumn{2}{c}{T=5} & \multicolumn{2}{c}{T=10} & \multicolumn{2}{c}{T=20} & \multicolumn{2}{c}{T=5} & \multicolumn{2}{c}{T=10} & \multicolumn{2}{c}{T=20}\\ 
\cmidrule(r){2-3} \cmidrule(r){4-5} \cmidrule(r){6-7} \cmidrule(r){8-9} \cmidrule(r){10-11} \cmidrule(r){12-13}
& \multicolumn{1}{c}{$A_{last}$} & $A_{avg}$ & \multicolumn{1}{c}{$A_{last}$} & $A_{avg}$ & \multicolumn{1}{c}{$A_{last}$} &$A_{avg}$ & \multicolumn{1}{c}{$A_{last}$} & $A_{avg}$ & \multicolumn{1}{c}{$A_{last}$} & $A_{avg}$ & \multicolumn{1}{c}{$A_{last}$} &$A_{avg}$\\
\midrule
EWC & \multicolumn{1}{c}{} 9.74& 19.92& \multicolumn{1}{c}{} 7.00& 15.35& \multicolumn{1}{c}{} 4.56& 12.51& \multicolumn{1}{c}{} 20.15& 32.10& \multicolumn{1}{c}{} 11.42& 23.52& \multicolumn{1}{c}{} 5.34& 15.58\\
online EWC & \multicolumn{1}{c}{}27.02& 39.63& \multicolumn{1}{c}{} 20.03& 31.76& \multicolumn{1}{c}{} 18.52& 31.96& \multicolumn{1}{c}{} 24.01& 35.41& \multicolumn{1}{c}{} 16.77& 27.63& \multicolumn{1}{c}{} 10.95& 19.44\\
SI & \multicolumn{1}{c}{} 7.46& 17.59& \multicolumn{1}{c}{} 4.04& 10.10& \multicolumn{1}{c}{} 2.24& 5.68& \multicolumn{1}{c}{} 12.14& 27.02& \multicolumn{1}{c}{} 6.97& 19.00& \multicolumn{1}{c}{} 3.81& 12.41\\
MAS & \multicolumn{1}{c}{} 25.53& 38.45& \multicolumn{1}{c}{} 15.32& 27.97& \multicolumn{1}{c}{} 11.67& 25.98& \multicolumn{1}{c}{} 20.04& 32.95& \multicolumn{1}{c}{} 11.56& 24.63& \multicolumn{1}{c}{} 7.41& 17.51\\
EWC-DR & \multicolumn{1}{c}{} \textbf{38.24}& \textbf{47.00}& \multicolumn{1}{c}{} \textbf{31.43}& \textbf{42.88}& \multicolumn{1}{c}{} \textbf{23.64}& \textbf{37.56}& \multicolumn{1}{c}{} \textbf{28.67}& \textbf{39.52}& \multicolumn{1}{c}{} \textbf{21.46}& \textbf{32.79}& \multicolumn{1}{c}{} \textbf{12.09}& \textbf{22.62}\\
\bottomrule
\end{tabular}
\label{tab:efcil_tinyim}
\end{table*}

\subsection{Exemplar-Free Class-Incremental Learning}
\label{sec:efcil}
\subsubsection{EFCIL Datasets} We utilize three public datasets: (1) CIFAR-100~\cite{krizhevsky2009learning}, 100 classes, 32$\times$32 pixels images, each class contains 600 images, with 500 allocated for training and 100 for testing; (2) ImageNet-Subset, 100 classes subset of ImageNet LSVRC dataset~\cite{russakovsky2015imagenet}, 224$\times$224 pixel images, each class consists of 1,300 training images and 50 validation images; (3) Tiny-ImageNet~\cite{le2015tiny}, a subset of ImageNet with 200 classes, 64$\times$64 pixel images, each class has 500 training images and 50 test images.

\subsubsection{Big-start Incremental Setting}
We adopt the big-start incremental learning protocol following~\cite{goswami2023fecam,petit2023fetril,zhu2021prototype,zhu2021class,zhu2022self}, where a substantial number of classes are allocated to the initial phase followed by evenly distributed incremental phases. The specific configurations for each dataset are as follows: CIFAR-100 and ImageNet-Subset are tested with (1) 50 initial classes and 5 incremental tasks of 10 classes, (2) 50 initial classes and 10 incremental tasks of 5 classes, (3) 40 initial classes and 20 incremental tasks of 3 classes. Tiny-ImageNet is tested with 100 initial classes and (1) 5 incremental tasks of 20 classes, (2) 10 incremental tasks of 10 classes, (3) 20 incremental tasks of 5 classes.

\subsubsection{Equally-split Incremental Setting} We also implement the equally split protocol described in~\cite{roy2024efficient,goswami2024resurrecting,smith2021always,gao2022r,meng2024diffclass}, where the entire set of classes is divided uniformly into 5, 10, or 20 tasks, with each task containing an equal number of classes. Notably, we choose the equally split setting because it poses a greater challenge for mitigating forgetting compared to the big-start setting, thereby providing a more rigorous evaluation.


\begin{figure}[t]
    \centering
    \includegraphics[width=0.95\linewidth]{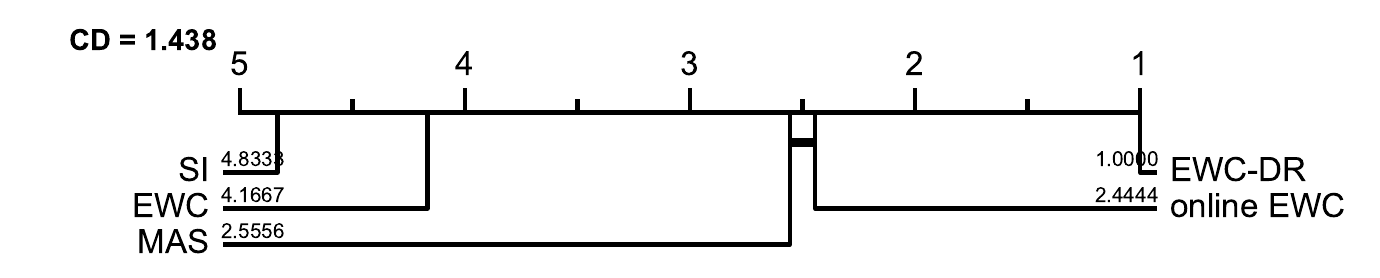}
    \caption{\lx{CD diagram comparing the $A_{avg}$ of different methods across all evaluated EFCIL settings. The CD is 1.438 with a significance level of 0.05.}}
    \label{fig:cd_diagram}
\end{figure}

\begin{figure*}[t]
\centering
\begin{subfigure}{0.28\linewidth}
  \centering
  \includegraphics[width=\linewidth]{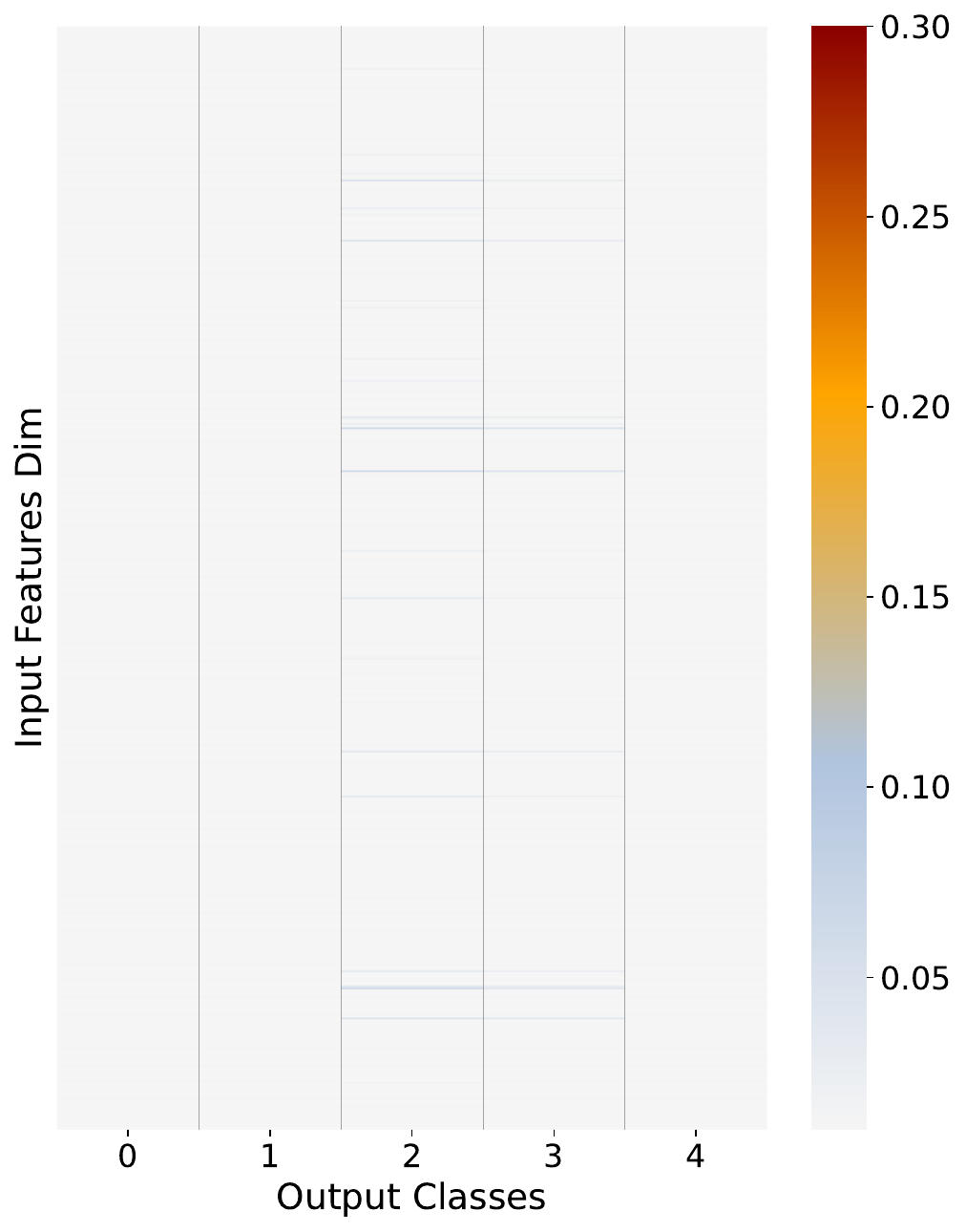}
  \caption{EWC}
  \label{fig:subfigavg2}
\end{subfigure}\hspace{2mm}
\begin{subfigure}{0.28\linewidth}
  \centering
  \includegraphics[width=\linewidth]{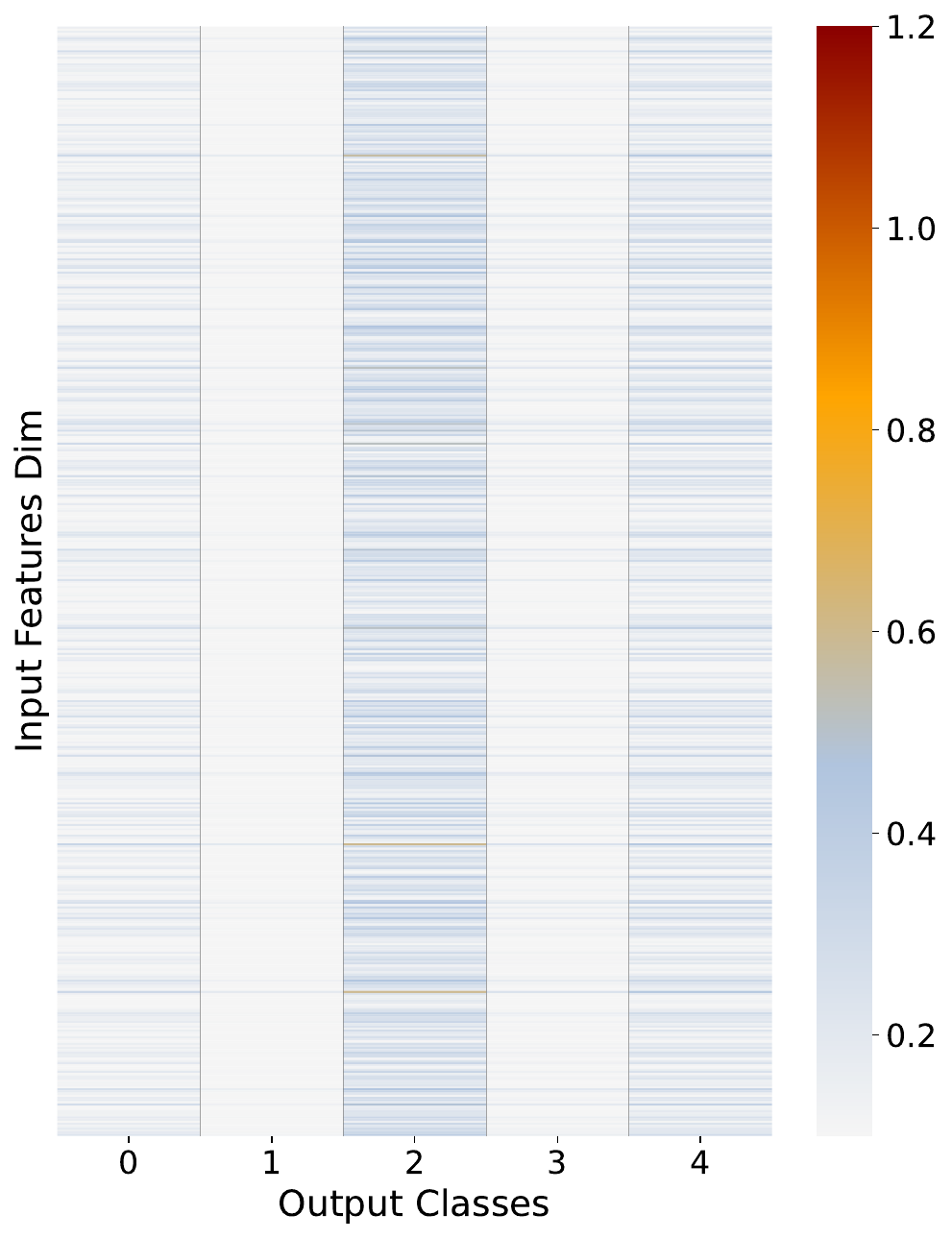}
  \caption{MAS}
  \label{fig:subfigavg3}
\end{subfigure}\hspace{2mm}
\begin{subfigure}{0.28\linewidth}
  \centering
  \includegraphics[width=\linewidth]{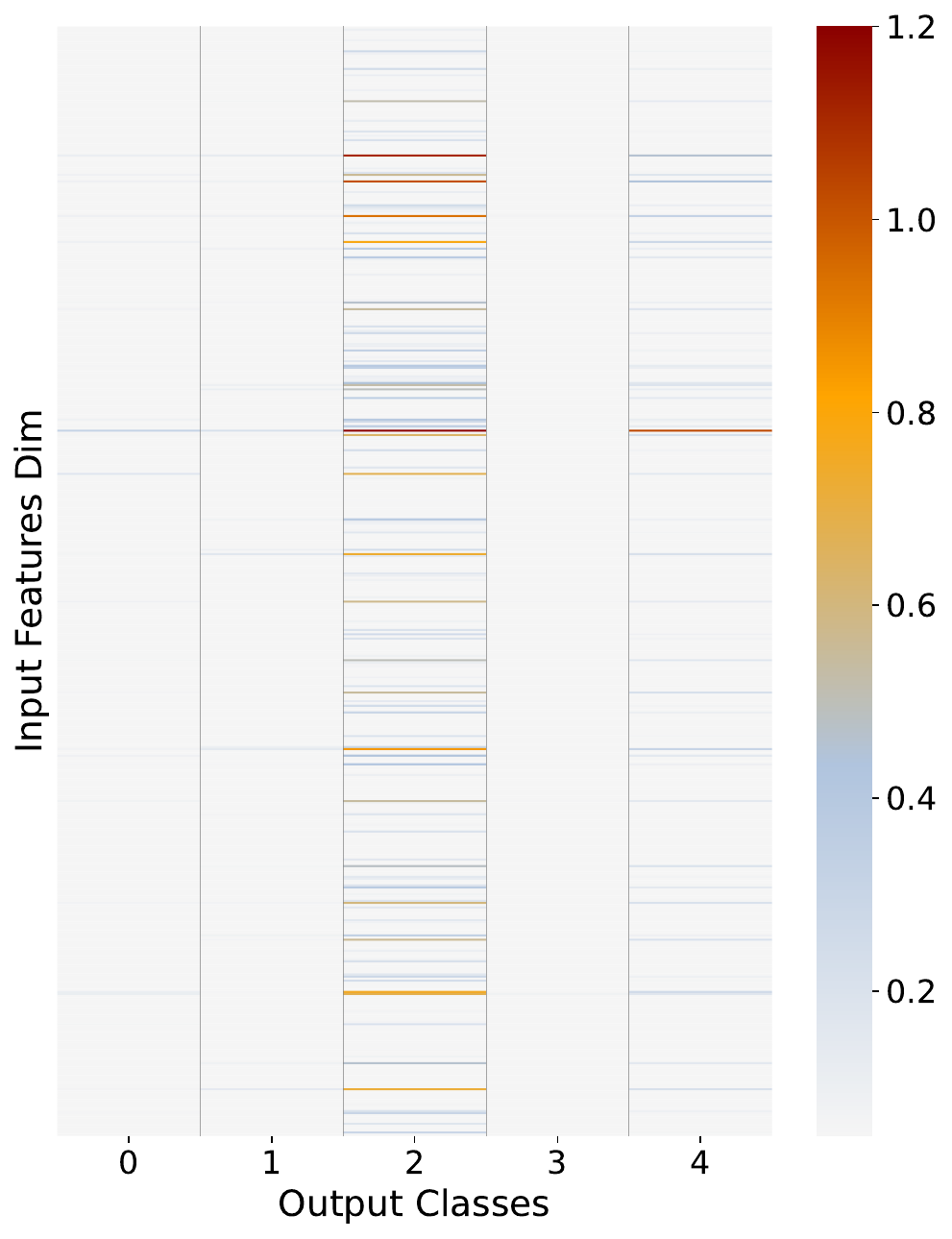}
  \caption{EWC-DR}
  \label{fig:subfigavg1}
\end{subfigure}
\caption{Importance \cxb{matrices} of the FC layer weights (ground truth at \cxb{class} 2) \cxb{for EWCs}.}
\label{fig:2d_avg_fcweight}
\end{figure*}

\subsubsection{\cxb{Implementation} Details} 
For all experiments, we adopt the PyCIL~\cite{zhou2023pycil} framework as our implementation basis. Consistent with prior work~\cite{rebuffi2017icarl,yan2021dynamically,zhu2021prototype,goswami2023fecam,meng2024diffclass,goswami2024resurrecting}, we employ a ResNet-18~\cite{he2016deep} model using SGD with an initial learning rate of 0.1, momentum of 0.9, batch size of 128, and weight decay of 0.0005. The model undergoes 200 training epochs with learning rate reductions by a factor of 10 at epochs 60, 120, and 160. For subsequent incremental tasks, we use a reduced initial learning rate of 0.1 with step-wise decay by a factor of 70 at epochs 120 and 150, and train for 180 epochs with a weight decay of 0.0002. All implementations are initialized and trained from scratch.  

\subsubsection{Evaluation \cxb{Metrics}}
Performance is evaluated using two metrics \cite{chaudhry2018riemannian,douillard2020podnet}: the last task accuracy ($A_{last}=A_{\mathcal{T}}$), which measures model performance after completing the final task $\mathcal{T}$, and the average incremental accuracy ($A_{avg}= \frac{1}{\mathcal{T}} \sum_{t=1}^\mathcal{T} A_t$), computed as the arithmetic mean of accuracies recorded after each task (including the initial task). Reported results are averaged across three independent runs.

\subsubsection{EFCIL Results}
As shown in~\cref{tab:efcil_cifar,tab:efcil_imsub,tab:efcil_tinyim}, our proposed EWC-DR consistently outperforms EWC and its variants across all evaluated settings. 
Specifically, EWC-DR achieves the best performance in both the big-start and equally split settings, achieving a maximum improvement of 53.18\% over EWC in $A_{last}$ and 55.47\% in $A_{avg}$. 
\lx{Online EWC outperforms EWC by accumulating importance weights online with a decay factor. However, its performance remains limited in the equally-split setting.}
Among the variants, MAS achieves the second-best results, possibly because MAS does not suffer from the gradient vanishing problem seen in EWC, SI and Online EWC. However, MAS is affected by redundant protection, which limits its overall effectiveness. 
More comparisons between our EWC-DR and CL methods from other categories can be found in the supplementary.

\lx{To assess the statistical significance of performance differences between methods across multiple datasets, we conduct a critical difference (CD) analysis on the experimental results under the EFCIL setting, as depicted in~\cref{fig:cd_diagram}. 
The CD diagram demonstrates that EWC-DR achieves the lowest average rank, indicating superior performance compared to all other methods. Notably, the gaps between EWC-DR and other methods exceed the CD threshold 1.438, confirming that the improvement of EWC-DR is statistically significant.
}

\subsubsection{\cxb{Statistics of FC Layer Weight Importance Matrices}}
\lx{To further support our main findings, we present the statistics on the importance of the fully connected (FC) layer weights.}
\cxb{The weight importance is represented as a matrix calculated using all CIFAR-100 training samples labeled as class 2 (GT) and averaged.}
\lx{\cref{fig:2d_avg_fcweight} visualizes the weight importance matrices of EWC, MAS, and EWC-DR, where warmer colors indicate higher importance values.}
\cxb{Our method emphasizes the ground truth class, yielding a prominent column for class~2.} 
\lx{In contrast, the heatmap \cxb{of} EWC appears nearly blank, reflecting \cxb{subtle} importance values due to gradient vanishing.}
\cxb{MAS highlights the ground truth class but also assigns high importance to class 0 and class 4, suggesting unnecessary protection.}
\cxb{Additionally, the importance value column of MAS at each activated class seems dense and evenly distributed, while that of EWC-DR appears more selective and concentrated.}
\cxb{Overall, our approach achieves a more selective and discriminative estimation of weight importance, mitigating both under-protection (as seen in EWC) and over-protection (as seen in MAS).
}



\subsection{Multimodal Continual Instruction Tuning}
\label{dec::mcit}

\begin{table*}[t]
\caption{
Results of MCIT. \lx{\cxb{The accuracy} $a_{t,j} \uparrow$ (\%) with forgetting transfer $\mathcal{F}_{t,j} \downarrow$ (\%) in brackets \cxb{are reported}.}
}
\centering
\begin{tabular}{lcccc}
\toprule
\multicolumn{5}{c}{\textbf{EWC}} \\
\midrule
\textbf{Checkpoint} & \textbf{VQA2} & \textbf{NLVR2} & \textbf{SNLI-VE}& \textbf{VCR}\\
\midrule
Task 1: VQA2                       & 67.89 & -- & -- & -- \\
Task 2: NLVR2                       & 39.94 [41.17] & 72.85 & --& --\\
Task 3: SNLI-VE                     & 40.83 [39.86] & 64.95 [34.59] &74.60 & -- \\
Task 4: VCR                         & 18.70 [72.46] & 52.13 [\emph{90.66}] & 41.84 [79.39] & 59.30\\
\bottomrule
\end{tabular}

\begin{tabular}{lcccc}
\toprule
\multicolumn{5}{c}{\textbf{EWC-DR}} \\
\midrule
\textbf{Checkpoint} & \textbf{VQA2} & \textbf{NLVR2} & \textbf{SNLI-VE}& \textbf{VCR}\\
\midrule
Task 1: VQA2                       & 67.72 & -- & -- & -- \\
Task 2: NLVR2                       & 43.25 [36.15] & 72.77 & --& --\\
Task 3: SNLI-VE                     & 54.25 [19.90] & 70.65 [9.31] &74.58 & -- \\
Task 4: VCR                         & 33.61 [50.37] & 66.51 [\emph{27.48}] & 50.86 [57.52] & 59.38\\
\bottomrule
\end{tabular}
\label{tab:mcit_full}
\end{table*}

\subsubsection{MCIT Datasets} 
\cxb{We follow CLiMB~\cite{srinivasan2022climb} benchmark and use four vision-and-language datasets:}
VQAv2~\cite{goyal2017making} is an image question answering dataset with short, open-ended answers. NLVR2~\cite{suhr2018corpus} is a language grounding dataset requiring reasoning about statements and paired photographs. SNLI-VE~\cite{xie2019visual} extends natural language inference to image-text pairs. VCR~\cite{zellers2019recognition} requires visual commonsense reasoning with multiple-choice questions.

\subsubsection{\cxb{Implementation} Details} 
Following the implementation in~\cite{srinivasan2022climb}, we employ a pre-trained Vision-Language Transformer (ViLT)~\cite{kim2021vilt} as the backbone encoder. The tasks are trained sequentially in the following order: VQAv2, NLVR2, SNLI-VE, and VCR. All training procedures and hyperparameters are kept consistent with those described in~\cite{srinivasan2022climb}.

\subsubsection{Evaluation \cxb{Metrics}}

\cxb{$a_{t,j}$ is defined as the accuracy on task $j$ evaluated after finishing training on task $t$. $A_{new} = a_{t,t}$ is reported to quantify the model plasticity.
The incremental accuracy $A_t$, evaluated on all seen classes right after being trained on task $t$, is also reported.}
\lx{Following~\cite{srinivasan2022climb}, we report forgetting transfer $\mathcal{F}_{t,j}$ , defined as $ \mathcal{F}_{t,j} = \frac{a_{j,j} - a_{t,j}}{a_{j,j} - a_{j-1,j}},\quad j < t $, where $a_{j-1,j}$ is the accuracy on task $j$ \cxb{before the learning task $j$}.}

\begin{figure}[t]
    \centering
    \includegraphics[width=0.85\linewidth]{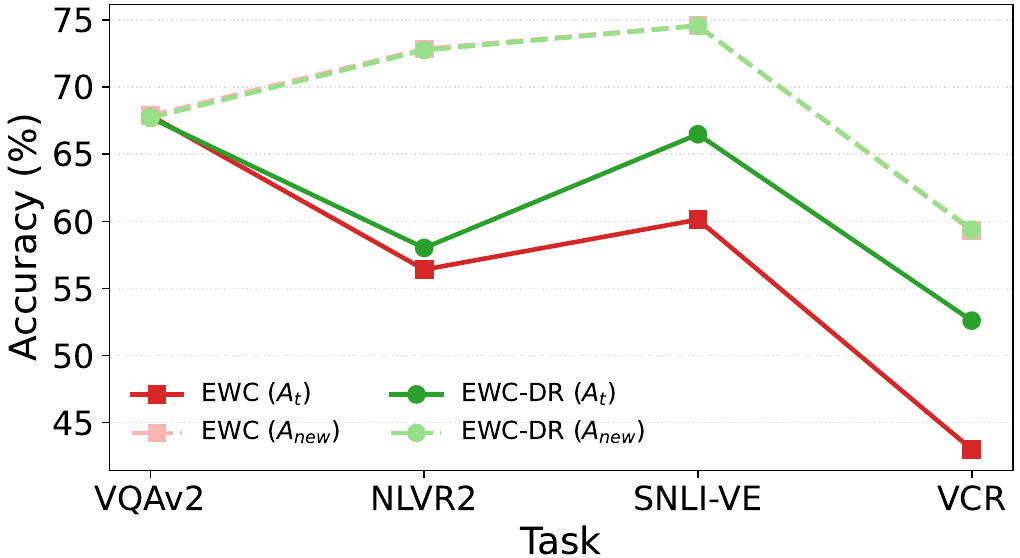}
    \caption{\cxb{The incremental accuracy} ($A_{t}$) and new task accuracy ($A_{\mathrm{new}}$) of EWC and EWC-DR across tasks in the MCIT scenario. Solid lines denote $A_{t}$, while dashed lines denote $A_{\mathrm{new}}$.}
    \label{fig:mcit}
\end{figure}

\subsubsection{MCIT Results}

\cxb{\cref{tab:mcit_full} presents the evaluation scores for each task after training on a subsequent task \(t\), along with the forgetting transfer values \(\mathcal{F}\) for all previously encountered tasks.
While vanilla EWC experiences catastrophic forgetting, e.g., a 90.66\% forgetting rate on NLVR2 after the final task, our EWC-DR effectively mitigates this issue, substantially reducing the corresponding forgetting rate to 27.48\%.
More importantly, such improvements are achieved without compromising the model plasticity, as EWC-DR achieves almost the same diagonal accuracies as those of EWC.}

\cxb{The comparisons between EWC and EWC-DR are illustrated in \cref{fig:mcit}.
Although both methods exhibit comparable new task learning ability (indicated by $A_{\text{new}}$), EWC-DR demonstrates significantly stronger resistance to forgetting, leading to improved performance on $A_t$. For instance, after training on the final VCR task, EWC-DR increases $A_{t}$ by nearly 10\%, from 42.99\% to 52.59\%, compared to EWC.
The analysis of \cref{fig:mcit} supports the findings in \cref{tab:mcit_full}.}


%% file: sec/6_conclusion.tex
\section{Conclusion}
In this work, we revisit Elastic Weight Consolidation (EWC) and its variants and diagnose their fundamental limitations for continual learning.
Through a novel gradient-based analysis framework, we find that EWC suffers from a vanishing gradient issue, leading to underestimated weight importance, while the EWC variant, Memory Aware Synapses (MAS), overly restricts irrelevant parameters, resulting in redundant protection.
To address these issues, we propose a simple operation, Logits Reversal (LR), to facilitate more accurate weight importance estimation.
Extensive experiments across multiple tasks and benchmarks demonstrate that EWC-DR consistently outperforms EWC and its variants. Therefore, we refer to this proposed method as EWC Done Right (EWC-DR).
The proposed analysis and method offer new insights for building more reliable and effective regularization-based continual learning algorithms.

\noindent\textbf{Acknowledgment}\quad 
This work was supported by the grants from the National Natural Science Foundation of China (62576369).

%% file: sec/X_suppl.tex
\clearpage
\setcounter{page}{1}
\maketitlesupplementary

\renewcommand{\thefootnote}{\fnsymbol{footnote}}
\renewcommand{\thefigure}{\Alph{figure}} 
\renewcommand{\thetable}{\Alph{table}} 
\renewcommand{\thelisting}{\Alph{listing}}  
\renewcommand{\thesection}{\Roman{section}}
\setcounter{figure}{0}
\setcounter{table}{0}
\setcounter{section}{0}
\setcounter{listing}{0}

\section{Pseudocode of Our Method}
We present the pseudocode of our proposed method in a PyTorch-like style in~\cref{lst:lr_code} to demonstrate its simplicity and effectiveness. After training the current network, we estimate the importance of each weight using Logits Reversal (LR) combined with cross-entropy loss. In contrast to EWC, which does not incorporate an LR step, our approach leverages this additional operation to achieve more accurate importance estimation.
\begin{listing}[!h]%
\caption{Pseudocode of Logits Reversal in PyTorch-like style.}%
\label{lst:lr_code}%
\begin{lstlisting}[language=Python]
"""
Input:
images: Tensor[bs, C, H, W]
targets: Tensor[bs,]

Output:
omega: dict{name: tensor, shape=param.shape}
"""
logits = network(images)["logits"]
logits_r = -logits
loss = torch.nn.functional.cross_entropy(logits_r, targets)
optimizer.zero_grad()
loss.backward()
for name, param in network.named_parameters():
    if param.grad is not None:
        omega[n]+= param.grad.pow(2).clone()  
\end{lstlisting}
\end{listing}

\section{Comparisons with Continual Learning Methods from Other Categories}
\footnotetext[2] {Replay-based method which stores and reuses samples from previous tasks during training.}
\begin{table}[ht]
\centering
\caption{EFCIL performance on CIFAR-100 compared against strong baselines. Reported results represent means across three independent trials.}
\begin{tabular}{ccccc}
\toprule
method & category & T=5 & T=10 & T=20 \\ 
\midrule
iCaRL-CNN$^\dagger$& Replay& 51.07 & 48.66 & 44.43 \\
iCaRL-NCM$^\dagger$& Replay& 58.56 & 54.19 & 50.51 \\
LWF-MC & Regularization & 45.93 & 27.43 & 20.07 \\
MUC & Regularization & 49.42 & 30.19 & 21.27 \\
GPM & Optimization & 41.51 & 37.78 & 41.27 \\
ADAM-NSCL & Optimization & 22.67 & 12.72 & 8.79 \\
EWC & Regularization & 32.82& 27.02& 19.61\\
\textbf{EWC-DR} & Regularization & \textbf{63.75}& \textbf{60.94}& \textbf{53.45}\\
\bottomrule
\end{tabular}
\label{tab:efcil_others}
\end{table}
In the CIFAR-100 big-start exemplar-free CIL setting, we report the average incremental accuracy ($A_{avg}$). Besides ours, a replay-based\footnotemark[2] method (iCaRL~\cite{rebuffi2017icarl} stores 20 samples per old class), two regularization-based methods (LWF-MC~\cite{rebuffi2017icarl}, MUC~\cite{liu2020more}), and two optimization-based ones (GPM~\cite{saha2021gradient}, ADAM-NSCL~\cite{wang2021training} reimplemented within our task-free framework) are included in~\cref{tab:efcil_others}. 
\lx{The EFCIL setting is significantly more challenging than replay-based ones since no previous task data can be stored or accessed during incremental learning.}

\begin{table}[h]
\centering
\caption{EFCIL results on CIFAR-100 comparing EWC-DR with modern methods. Average incremental accuracy ($A_{avg}$) is reported using seed 1993.}
\begin{tabular}{ccccccc}
\toprule
\multirow{2}{*}{method} & \multicolumn{3}{c}{Big start} & \multicolumn{3}{c}{Equally split} \\
\cmidrule(r){2-4} \cmidrule(r){5-7}
& T=5 & T=10 & T=20 & T=5 & T=10 & T=20 \\
\midrule
FeTrIL & 66.3 & 65.2 & 61.5 & 60.4 & 52.1 & 43.2 \\
PASS & 63.5 & 61.8 & 58.1 & 63.4 & 52.2 & 41.8 \\
SSRE &  65.9&  65.0& 61.7 & 56.6 & 44.4 & 33.6 \\
EWC &  33.4&  26.4& 19.2 & 39.2 & 27.3 & 17.7 \\
EWC-DR & 63.9 & 62.1 & 53.5 & 61.5 & 49.6 & 35.3 \\
\bottomrule
\end{tabular}
\label{tab:sota}
\end{table}
\cref{tab:sota} showcases the comparison of EWC-DR with other modern PyCIL methods on CIFAR-100 under seed 1993, our method substantially elevates the original EWC from a lower baseline to a level competitive with contemporary techniques. Notably, in the Equally-split setting, EWC-DR even surpasses the SSRE. 

Our proposed EWC-DR significantly improves upon EWC, making it a competitive baseline in continual learning especially in exemplar-free setting, though it is not intended to be state-of-the-art.

\section{Sensitivity Analysis of \texorpdfstring{$\lambda$}{lambda}}
\label{sec::hyperparam}
\begin{figure}[ht]
    \centering
    \includegraphics[width=0.75\linewidth]{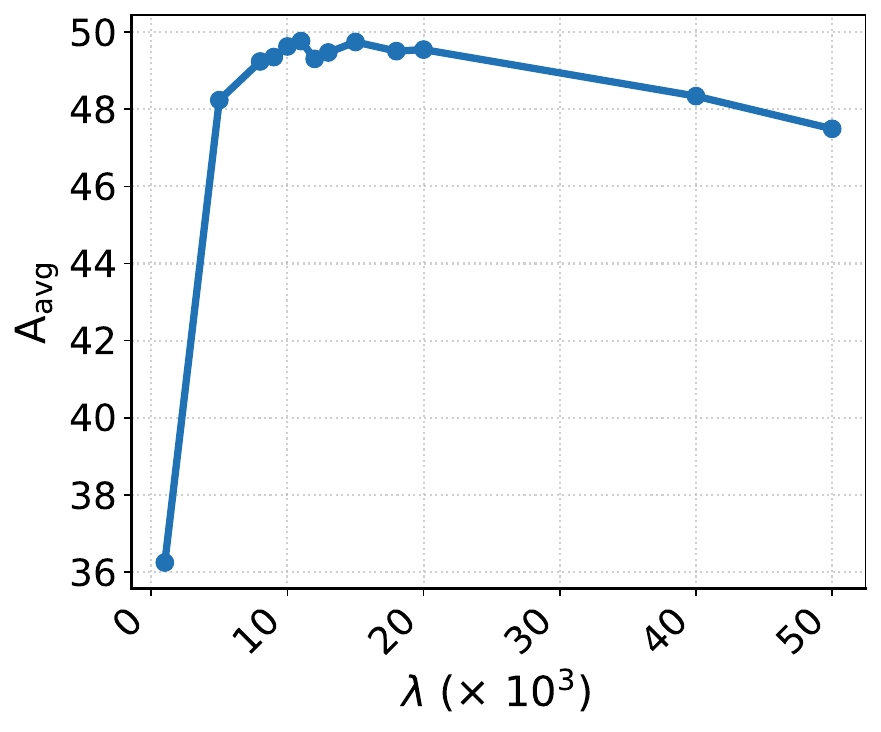}
    \caption{%
    Sensitivity analysis of the average accuracy ($A_{avg}$) with respect to the regularization parameter $\lambda$ on CIFAR-100 under the 10-task equally split EFCIL setting.
    }
    \label{fig:analysis_lambda}
\end{figure}
As with all weight regularization methods, our EWC-DR introduces a hyperparameter $\lambda$, which penalizes changes to the model weights, as defined in~\cref{ewc_regloss}. 
We conducted a sensitivity analysis to assess the impact of $\lambda$ on the average accuracy ($A_{avg}$). Specifically, for the CIFAR-100 10-task equally split EFCIL setting, $\lambda$ was varied from 1{,}000 to 50{,}000.
As depicted in~\cref{fig:analysis_lambda}, $A_{avg}$ increases steadily as $\lambda$ grows from 1{,}000 and reaches its peak around $\lambda = 10{,}000$--$20{,}000$. Beyond this point, further increasing $\lambda$ leads to a gradual decline in accuracy. 
This trend suggests that an appropriate choice of $\lambda$ is crucial for balancing the trade-off between learning new knowledge and mitigating forgetting.
However, the proposed method is not sensitive to $\lambda$, as shown in \cref{fig:analysis_lambda}.

\section{\cxb{Statistics of FC Layer Weight Importance Matrices}}
\cxb{We report statistics on the importance of the FC layer weights in tabular form, corresponding to the illustrations in \cref{fig:2d_avg_fcweight}. }
Importance values per class are \cxb{further} summed for tabulation in~\cref{tab:avg_fcweight}. 
EWC has low importance due to gradient vanishing. MAS has high importance and a relatively even distribution across classes. Our EWC-DR obtains high importance, with a notable peak in the GT class. 
\begin{table}[ht]
\caption{
Statistics of the FC layer weight importance.
}
\centering
\begin{tabular}{cccccc}
\toprule
class & 0 & 1 & 2 (GT) & 3 & 4 \\
\midrule
EWC & 0.003 & 0.013 & 0.219 & 0.136 & 0.018 \\
MAS & 14.44 & 8.53 & 25.48 & 10.26 & 18.08 \\
EWC-DR & 5.30 & 6.55 & 41.38 & 2.51 & 11.12 \\
\bottomrule
\end{tabular}
\label{tab:avg_fcweight}
\end{table}